\documentclass[sigconf]{acmart}

\usepackage{graphicx}
\usepackage{amsmath}
\usepackage[ruled,linesnumbered]{algorithm2e}
\usepackage{bm}
\usepackage{booktabs}
\usepackage[switch]{lineno}
\usepackage{tabularx}
\usepackage{microtype}

\usepackage{algorithm}
\usepackage{subfigure}
\usepackage{multirow}
\usepackage{hhline}
\usepackage{xcolor}
\usepackage{enumitem}

\usepackage{makecell}

\AtBeginDocument{%
  }

\renewcommand\footnotetextcopyrightpermission[1]{} 
\begin{document}


\title{SENSE: Satellite-based ENergy Synthesis for Sustainable Environment}

\author{Kailai Sun}
\orcid{0000-0003-1648-3409}
\affiliation{%
  \institution{SMART, Singapore}
  \country{ }
}
\affiliation{%
  \institution{Massachusetts Institute of Technology}
  \city{Cambridge}
  \state{MA}
  \country{USA}
}
\email{skl24@mit.edu}

\author{Mingyi He}
\affiliation{%
  \institution{Massachusetts Institute of Technology}
  \city{Cambridge}
  \state{MA}
  \country{USA}
}
\email{mingyihe@mit.edu}

\author{Heye Huang}
\affiliation{%
  \institution{SMART, Singapore}
  \country{}
}
\affiliation{%
  \institution{Massachusetts Institute of Technology}
  \city{Cambridge}
  \state{MA}
  \country{USA}
}
\email{heyeh@mit.edu}

\author{Can Rong}
\affiliation{%
  \institution{SMART, Singapore}
  \country{}
}
\affiliation{%
  \institution{Massachusetts Institute of Technology}
  \city{Cambridge}
  \state{MA}
  \country{USA}
}
\email{rongcan@mit.edu}

\author{Alok Prakash}

\affiliation{%
  \institution{SMART, Singapore}
  \country{}
}
\affiliation{%
  \institution{Massachusetts Institute of Technology}
  \city{Cambridge}
  \state{MA}
  \country{USA}
}
\email{alokprks@mit.edu}

\author{Baoshen Guo}
\affiliation{%
  \institution{SMART, Singapore}
  \country{}
}
\affiliation{%
  \institution{Massachusetts Institute of Technology}
  \city{Cambridge}
  \state{MA}
  \country{USA}
}
\email{baoshen@mit.edu}

\author{Shenhao Wang}
\orcid{0000-0003-4374-8193}
\affiliation{%
  \institution{University of Florida}
  \city{Gainesville}
  \state{Florida}
  \country{USA}}
\authornotemark[1]
\email{shenhaowang@ufl.edu}

\author{Jinhua Zhao}
\orcid{0000-0002-1929-7583}
\affiliation{%
  \institution{Massachusetts Institute of Technology}
  \city{Cambridge}
  \state{Massachusetts}
  \country{USA}}
\authornote{Jinhua Zhao and Shenhao Wang are the corresponding authors.}
\email{jinhua@mit.edu}

\renewcommand{\shortauthors}{Kailai et al.}

\begin{abstract}
Urban Building Energy Modeling (UBEM) plays a critical role in achieving the United Nations' Sustainable Development Goals 7 and 11. Although existing studies based on satellite imagery and deep learning have achieved remarkable progress, many challenges exist: most existing studies are inherently predictive, failing to reflect the generative nature of urban planning; although generative AI and diffusion models have seen explosive growth in satellite imagery, they lack the corresponding urban functional generation (e.g., energy layer); third, aligned high-quality high-resolution building energy data with satellite imagery is limited and scarce. To address them, we propose \textbf{SENSE} (Satellite-based ENergy Synthesis for Sustainable Environment), a unified generative UBEM framework that jointly synthesizes realistic urban satellite imagery and aligned high-quality building energy consumption and height maps. By conditioning on road networks and urban density metrics, our framework, based on a controllable diffusion model, leverages the knowledge learned by large vision models to generate urban building energy consumption and height information (annotations) in the latent space. Experiments across four cities (New York City, Boston, Lyon, and Busan) demonstrate that SENSE achieves high visual fidelity and strong physical consistency, satisfying the ASHRAE standard. Experiments demonstrate that SENSE can generate enough annotated synthetic data using less than 20\% labeled energy data, boosting downstream prediction performance by 10\% IoU.  Compared to state-of-the-art urban building energy prediction methods, SENSE significantly reduced prediction error (reduced 3\%-11\% NMBE and 1\%-9\% CVRMSE). This study offers an energy-efficiency urban planning and physical generation solution for urban science, energy science and building science. The dataset and code links: \url{https://huggingface.co/datasets/skl24/MUSE} and \url{https://github.com/kailaisun/GenAI4Urban-Energy/}.
\end{abstract}





\maketitle

\section{Introduction}
Urban residents comprise 55\% of the global population, a figure projected to rise to 68\% by 2050 \cite{UN2018Urbanization}. The rapid urbanization of the global population has positioned cities as the primary battleground for climate change mitigation, and nearly 70 \% of world energy is consumed by urban activities \cite{dai2025citytft}. Buildings are a major contributor to global energy consumption and greenhouse gas emissions, accounting for 32 per cent of global energy demand and 34 per cent of CO$_2$ emissions \cite{GlobalABC2024}. The total building energy consumption mainly includes Heating, Ventilating and Air-Conditioning (HVAC) and lighting systems \cite{sun2020review}. The global imperative to decarbonise cities has placed Urban Building Energy Modeling (UBEM) at the forefront of sustainable development research. Effective modeling and planning of urban energy dynamics is essential for policy-making and achieving United Nations' (UNs') Sustainable Development Goals (SDGs), specifically SDG 7 (Affordable and Clean Energy) and SDG 11 (Sustainable Cities and Communities). By optimizing the urban and building designs and improving the energy efficiency,  this domain can make a significant contribution to creating a high-quality and low-emission built environment \cite{zhou2025state}.

Existing studies usually use satellite imagery as an essential tool for urban monitoring, evaluation, and prediction, because satellite imagery provides rich information. \citet{WANG2025106054} utilized Mask-RCNN to extract 2.5D building massing and type from satellite imagery for urban building energy modelling in Chicago and San Francisco. \citet{streltsov2020estimating} train CNNs to segment and predict residential building energy consumption at the building level using overhead imagery. \citet{YANG2025115522} use GCN-LSTM model to perform spatiotemporal predictions of urban building rooftop photovoltaic potential with satellite imagery.  \citet{WANG2025119218} proposed a satellite image encoder with spatio-temporal vision transformer and multi-modal fusion to predict urban power. \citet{MAYER2023120542} and \citet{streltsov2020estimating} apply aerial imagery and street view imagery to estimate building energy efficiency using computer vision models (e.g., Resnet and Inception). \citet{FEHRER2018252} use nighttime light images \cite{wang2024estimation} to explain upwards of 90\% of the variability in energy consumption in the United States. Recently, with the development of GenAI, \citet{WANG2025102339} use diffusion models to generate high-fidelity satellite imagery for automating urban planning in Chicago, Dallas, and Los Angeles. \citet{HE2026113892} apply multi-stage diffusion models to generate building layouts and satellite imagery for urban planning in Chicago and New York City (NYC). 

On the other hand, traditional physics-based urban and building energy simulation approaches, often calculate thermal dynamics based on detailed building physics and meteorological inputs \cite{reinhart2016urban}.  \citet{bian2025integrating} proposed an integrated workflow coupling microclimate modelling (ENVI-met) with energy simulation to capture the feedback loops between urban morphology and local thermal environments. \citet{li2025integrating} emphasized the necessity of integrating Environmental Impact Assessment (UB-EIA) into energy modeling to evaluate the lifecycle carbon footprint of urban developments. Beyond physics-based studies, data-driven studies \cite{ali2023review} become hot topics. Authors \citet{dai2025citytft} introduced CityTFT, a Temporal Fusion Transformer-based model that predicts heating and cooling loads up to 240 times faster than traditional physics engines.

With the rapid development of computer vision and remote sensing \cite{10179972,zhao2024artificial}, GenAI and diffusion methods \cite{ho2020denoising} have become mainstream. CRS-Diff \cite{tang2024crsdiff} introduced controllable satellite imagery generation to remote sensing, by integrating text prompts, metadata, and segmentation maps. Diffusionsat \cite{khanna2024diffusionsat} proposed a generative foundation model from Stable Diffusion (SD) and latent variants (LDMs) \cite{rombach2022high} for satellite imagery generation using remote sensing metadata.  \citet{11223686} proposed a dual loop data cleaning method to generate high-quality data for remote sensing generation models. 
 
Although existing studies have achieved remarkable progress, existing UBEM studies are constrained by fundamental methodological and data challenges. First, most existing UBEM studies are inherently predictive (e.g., they map input geometry, image and weather to predict energy consumption). They can evaluate and predict metrics from a given urban plan, but it is hard to generate new, energy-efficient urban morphologies. Second, although diffusion models have seen explosive growth in satellite imagery, these models operate primarily in the visual domain (RGB). They lack the corresponding urban functional generation (e.g., energy layer) in the urban field. Third, developing accurate data-driven UBEMs requires large datasets of aligned satellite imagery and high-quality building energy records. However, such data is scarce and sparse due to privacy, cost, sensitivity, etc \cite{ali2023review}. Deep learning models trained on limited data usually overfit and fail to generalize across different real-world scenes.

To address these challenges, in this study, we propose a unified multi-modal generative AI framework for both urban satellite imagery and building energy generation. By conditioning on road networks and text-based urban density metrics, our framework can simultaneously generate realistic and diverse urban satellite imagery, aligned and corresponding high-quality building energy consumption and height maps. Our framework is a controllable diffusion model conditioned on road networks and urban density metrics, integrated with the proposed building energy decoder and height decoder. Because existing large GenAI computer vision models can implicitly learn rich visual representations, we leverage the knowledge learned by these models to generate urban building energy consumption and height information in latent space, instead of training a joint generator from scratch. We validate our framework on a multi-city global dataset covering New York City, Boston, Lyon, and Busan.
The main contributions are:


\begin{itemize}
    \item We propose the unified multi-modal GenAI framework that generates satellite imagery and corresponding urban building energy consumption and height maps, conditioned on road-network constraints and urban density metrics.
    
    \item By extending the co-generated urban modalities (e.g., energy and height decoders with 89.25\% and 85.75\%accuracies), we demonstrate that urban building energy consumption (achieves NMBE of 3.05\% and CVRMSE of 14.62\%) and height can be reliably generated from the latent space. 

    \item We establish a global Multi-city Urban Satellite-Energy Dataset(MUSE) covering NYC, Boston, Lyon, and Busan, where municipal-scale energy disclosure records are spatially aligned with high-resolution satellite imagery.
    
    \item For the energy data scarcity issue, experiments demonstrate that our generative data augmentation strategy with limited real data (less than 20\%) improves the performance of energy prediction models by 10\% mIoU.  Compared to existing urban building energy prediction methods, our strategy significantly reduced energy prediction error (reduced 3\%-11\% NMBE and 1\%-9\% CVRMSE).
\end{itemize}

\vspace{-1em}
\section{Multi-city Urban Satellite-Energy Dataset}


\subsection{Data Coverage}

\begin{figure*}[!ht]
    \centering
    \vspace{-1em}
    \includegraphics[width=0.88\linewidth]{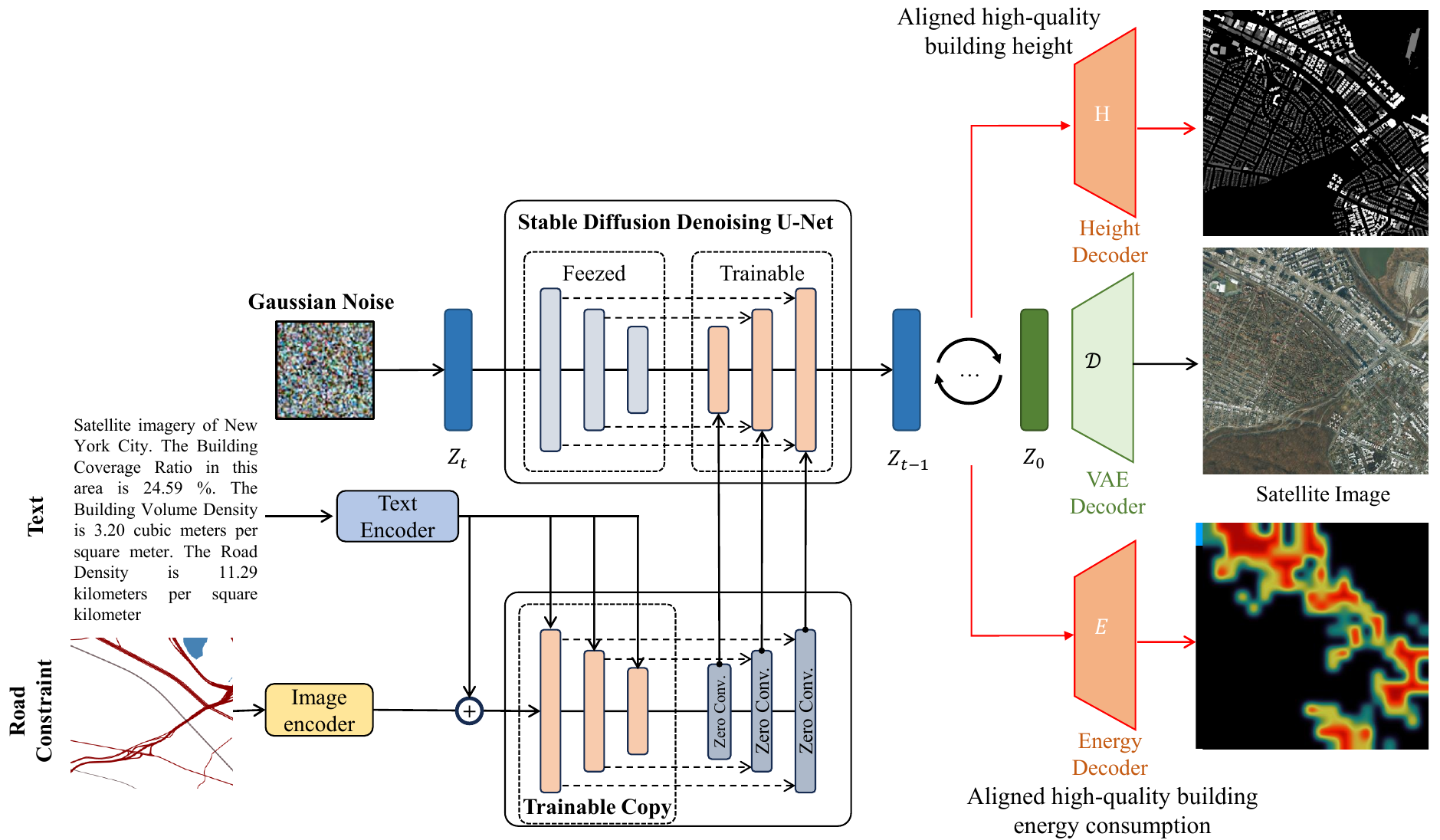}
    \vspace{-1em}
    \caption{Proposed GenAI framework for generating satellite image,  building height and building energy consumption together.}
    \label{fig:framework}
    \vspace{-1em}
\end{figure*}

We established a new global multi-city dataset, as defined by the GHS Urban Centre Database \cite{maririvero2024ghsurban}, spanning four cities: North America (NYC and Boston), Western Europe (Lyon), and East Asia (Busan). We align municipal-scale building energy disclosure records with satellite imagery and create paired samples at a fixed spatial extent in Tab. \ref{tab:city_dataset_stats} in Appendix section~\ref{data-filter}. Specifically, in Fig. \ref{fig:framework}, each sample corresponds to a $2\,\mathrm{km}\times 2\,\mathrm{km}$ tile, represented by (1) an urban satellite image, (2) a text prompt with urban density metrics, (3) a geospatial constraint map with water, railway and main roads, (4) a building-level height map and (5) a building-level energy map where the energy values transformed by a log1p function. 

\vspace{-1em}
\subsection{Dataset Overview}


\subsubsection{Satellite Imagery and Urban Density}
Urban boundary data are obtained from the Global Human Settlement (GHS) Urban Centre Database 2023~\cite{ghs_ucdb}. High-resolution satellite imagery was obtained from Mapbox~\cite{mapbox_api}, then cropped and mosaicked into $512 \times 512$ pixel tiles aligned with each grid. Building attributes were derived from the Global Human Settlement Layer (GHSL P2023A)~\cite{pesaresi2024advances}. For each $2\text{ km} \times 2\text{ km}$ cell, we computed three density metrics: (1) Building Volume Density (BVD) = total built-up volume / land area; (2) Building Coverage Ratio (BCR) = total built-up area / land area; and (3) Road Density (RD)= total road area / land area.

\subsubsection{Geospatial Constraint}
We derive geospatial constraints from OpenStreetMap (OSM)~\cite{openstreetmap}, a public database of vectorized urban features. We specifically extract water bodies, railway infrastructure, and major road networks (ranging from motorways to tertiary roads). Minor streets are intentionally excluded to avoid over-constraining the generation of local details. Technically, we perform a spatial intersection between these vectorized layers and each target grid cell, subsequently rasterizing the outputs into $512 \times 512$ pixels binary masks to serve as the spatial control conditions.

\subsubsection{Building Height and Footprint}
To construct accurate 3D urban morphological ground truth, we primarily leveraged the 3D-GloBFP dataset~\cite{che20243d}, which serves as the global open-source 3D building footprint database. To ensure the highest fidelity for our target cities, we cross-referenced and supplemented this with local high-resolution authoritative data. Specifically, for the NYC (NYC) case study, building footprints and height attributes were extracted from the official NYC Department of City Planning database~\cite{nyc_buildings}. For the Lyon case study, we utilized the 3D city model data~\cite{che_2024_11391077}, which provides detailed height information. These datasets were rasterised to match the spatial resolution ($512 \times 512$ pixels).

\subsubsection{Building Energy Consumption}

High-quality ground truth is important for training the energy decoder. We compiled a multi-source dataset of annual building energy consumption in 2023, using municipal disclosure records. For NYC, we leveraged the Energy and Water Data Disclosure (Local Law 84) dataset~\cite{nyc_ll84}, which mandates buildings energy benchmarking; for Boston, the Building Emissions Reduction and Disclosure Ordinance (BERDO) registry~\cite{boston_berdo}; for Lyon, address-level energy consumption records from the Metropolis of Lyon via the French national open data platform~\cite{lyon_energy}; for Busan, the Busan Metropolitan City administrative database~\cite{busan_city}. Because the energy data (kBtu) exhibit a long-tailed distribution, we apply the log1p function to transform the data into a Gaussian-like distribution.


\vspace{-1em}
\subsection{Data Pre-processing}

We perform the spatial alignment across high-resolution satellite imagery, geospatial constraints, building height and energy disclosure records, ensuring precise synchronization through a unified geodetic coordinate system. Because MUSE is established by spatially aligning heterogeneous sources, we apply tile-level quality control to remove samples with unreliable and missing building energy annotations (Appendix section~\ref{data-filter} in Fig. \ref{fig:datafilter}).

We recruited three urban domain specialists to manually review the energy annotations, flagging a tile as unaccepted if its energy label map exhibited large contiguous blocks of missing or null values. We use expert-in-the-loop filtration to ensure that the model is trained on high-quality samples where the spatial distribution of the energy label map aligns with the observed urban morphology. Finally, the high-quality dataset comprises 2,788 tiles in total, including 579 tiles for NYC, 526 for Boston, 687 for Lyon, and 996 for Busan, providing a data foundation for subsequent analysis.  To facilitate further scientific research in urban and energy domains and ensure the reproducibility, we have publicly released the full MUSE dataset at the Hugging Face: \url{https://huggingface.co/datasets/skl24/MUSE}. We encourage the community to benchmark and extend GenAI applications for urban and energy sustainability across cities.



\vspace{-1em}

\section{Method}
\label{sec:methodology}

We propose a unified multimodal generative AI framework to generate realistic and controllable urban satellite imagery, high-quality building energy consumption and building height maps together, conditioned on textual and spatial inputs, such as urban density metrics and road networks. In particular, our framework aims to model the joint distribution $P(\mathbf{x}, \mathbf{y}_{e}, \mathbf{y}_{h} | \mathbf{c})$ of satellite imagery $\mathbf{x}$, building energy consumption maps $\mathbf{y}_{e}$, and building height maps $\mathbf{y}_{h}$, conditioned on urban constraints $\mathbf{c}$. In Fig \ref{fig:framework}, our framework decouples the generation process into two stages: (1) we train a controllable latent diffusion model to obtain the visual latent feature; and (2) we train building decoders (building height and energy) to extract height and energy layers in the latent space.

\subsection{Controllable Geospatial Diffusion Model}
The foundation of our framework is the generation of realistic and diverse urban imagery that conditions on natural language (e.g., by prompting for variations in urban density) with strict geospatial constraints (e.g., road networks). To achieve this, we leverage Latent Diffusion Models (LDMs) \cite{rombach2022high} augmented with ControlNet \cite{zhang2023adding}.

\subsubsection{Preliminaries on latent diffusion models}
A pre-trained Variational Autoencoder (VAE) consists of an encoder $\mathcal{E}$ and a decoder $\mathcal{D}$. Given a real satellite image $\mathbf{x} \in \mathbb{R}^{H \times W \times 3}$, the encoder maps it to a latent representation $\mathbf{z}_0 = \mathcal{E}(\mathbf{x}) \in \mathbb{R}^{h \times w \times c}$. The diffusion process is modeled as a forward Markov chain that progressively adds Gaussian noise to $\mathbf{z}_0$ over $T$ timesteps, producing a sequence $\mathbf{z}_1, \dots, \mathbf{z}_T$. The reverse process aims to recover $\mathbf{z}_0$ from noise $\mathbf{z}_T \sim \mathcal{N}(0, \mathbf{I})$ via a denoising U-Net $\epsilon_\theta$. The optimization objective is to minimize the noise prediction error:
\begin{equation}
    \mathcal{L}_{LDM} = \mathbb{E}_{\mathbf{z}_0, t, \mathbf{c}_{txt}, \epsilon \sim \mathcal{N}(0, 1)} \left[ \| \epsilon - \epsilon_\theta(\mathbf{z}_t, t, \mathbf{c}_{txt}) \|_2^2 \right],
\end{equation}
where $t$ is the time step, and $\mathbf{c}_{txt}$ represents the text condition (e.g., "Satellite imagery of New York City. The Building Coverage Ratio in this area is 24.59 \%. The Building Volume Density is 3.20 cubic meters per square meter. The Road Density is 11.29 kilometers per square kilometer").

\subsubsection{Geospatial environmental constraints}
Text-to-image generation models often hallucinate buildings in physically invalid locations. To ensure morphological consistency, we introduce a geospatial environmental constraint $\mathbf{c}_{env}$ using ControlNet. We first create a trainable copy of the encoding blocks of the Stable Diffusion encoder. Then, let $\mathcal{F}(\cdot; \Theta)$ denote a neural network block with weights $\Theta$. "Zero convolution" layers $\mathcal{Z}$ are initialized with zeros and weights $\Theta_{copy}$. The output of a controlled block $\mathbf{y}$ is:
\begin{equation}
    \mathbf{y} =  \mathcal{F}(\mathbf{x}_{in} + \mathcal{Z}(\mathbf{c}_{env}); \Theta_{copy}) ).
\end{equation}
$\mathcal{Z}(\mathbf{c}_{env})$ does not influence the base model at the start of training, preserving the pre-trained visual knowledge. As training progresses, it learns to inject the geospatial environmental information $\mathbf{c}_{env}$ into the feature space, ensuring that the generated urban imagery strictly respects the topological boundaries. The geospatial environmental constraints (e.g., road network, water, etc.) are important for accurate urban energy modeling.

\subsection{Energy and Height Decoders}
While the diffusion model can generate the visual urban imagery (RGB), existing studies have not considered the co-generation of building height and building energy. A core hypothesis of this study lies in that the high-level semantic features required to generate a realistic urban imagery (e.g., residential buildings, factories) are intrinsically correlated with building height and energy. Instead of training separate generative models for each modality from scratch, we use the weights of the visual generation module and add lightweight “plug-and-play” decoders to extract specific building height and energy features in the latent space.

\subsubsection{Multi-Scale Feature Extraction}
Let $\Psi_{SD}$ be the U-Net of the diffusion model. During the denoising process at a fixed timestep $t^*$, we extract a set of hierarchical feature maps $\{\mathbf{f}_i\}_{i=1}^{K}$ from the decoder blocks of the U-Net. These features contain rich semantic information at different resolutions (e.g., $64\times64$).
\begin{equation}
    \mathbf{F}_{latent} = \text{Concat}\left( \text{Upsample}(\mathbf{f}_1), \dots, \text{Upsample}(\mathbf{f}_K) \right).
\end{equation}
$\mathbf{F}_{latent}$ serve as the shared representation for all decoders.

\subsubsection{H-Decoder}

To recover the 3D structure of the generated city, we design the Height-Decoder (H-Decoder) to generate building height levels. Instead of continuous regression, we formulate this as a generative segmentation task to handle the discrete urban data. We employ the SegFormer architecture equipped with Mix Transformer (MiT) encoders to capture multi-scale latent features.

We discretize the spatial data into $N_h=5$ distinct categories, where Class 0 represents non-building background areas, and Classes 1--4 represent increasing building height intervals. The H-Decoder outputs a probability map $\hat{\mathbf{Y}}_{h} \in \mathbb{R}^{H \times W \times 5}$, learning distinct morphological patterns associated with different building height tiers (e.g., low-rise residential and high-rise commercial). The loss function follows the standard segmentation formulation, combining Cross-Entropy loss and Dice loss to ensure pixel-level accuracy and region-level consistency.

\subsubsection{E-Decoder}
The challenging task is generating the building energy consumption. We consider that visual features encoded in the latent space $\mathbf{F}_{latent}$ (e.g., roof size, texture, building density) can be leveraged for physical energy generation.

Similar to the H-Decoder, we discretize the continuous energy consumption values into $N_e=4$ classes: Class 0 denotes non-energy areas (background), and Classes 1-3 correspond to Low, Medium, and High energy consumption levels, respectively. To address the inherent class imbalance in energy data (where high-consumption buildings are rare), we implement a class-weighted cross-entropy loss combined with Dice loss:
\vspace{-0.5em}
\begin{equation}
    \mathcal{L}_{energy} = -\sum_{c=0}^{N_e-1} \omega_c \mathbf{Y}_{e,c} \log(\hat{\mathbf{Y}}_{e,c}) + \mathcal{L}_{Dice}(\hat{\mathbf{Y}}_{e}, \mathbf{Y}_{e}),
\end{equation}
where $\omega_c$ represents the weight assigned to class $c$ (calculated inversely proportional to class frequency) to penalize errors on minority classes (e.g., buildings with high-energy consumption).



\section{Experiments}
In this section, we conduct extensive experiments to answer the following research questions:
\begin{itemize}[leftmargin=*]
    \item RQ1: How effectively does the proposed framework generate physically consistent and spatially aligned building height/energy consumption maps with generated urban imagery? 
    \item RQ2: To what extent do the generated physical energy consumption data align with established industry standards for UBEM? 
    \item RQ3: Can the knowledge in existing GenAI methods be leveraged for accurate urban physical prediction with limited real data?
    \item RQ4: How does our framework improve existing state-of-the-art urban building energy prediction models with limited real data?
\end{itemize}

\vspace{-1em}
\subsection{Experimental Setups}
Experiments were conducted on a 64-bit Linux 22.04 platform equipped with an Intel(R) Xeon(R) Gold 6438N 128-core processor, 500 GB RAM, and 8 NVIDIA H100 GPUs (80 GB memory each) running CUDA 12.0. The programming environment was Python 3.10 and PyTorch 1.13. 
For implementation details and hyperparameters, please see the Appendix section~\ref{sec:Implementation}.

\subsection{Energy Metric}

While semantic segmentation metrics (e.g., mIoU) often evaluate pixel-level segmentation accuracy, urban energy planners are often concerned with the reliability of total energy demand estimates at the district scale. Following ASHRAE Guideline 14 \cite{american2014ashrae,ROYAPOOR2015109}, we adopt industry-standard calibration metrics: the Normalized Mean Bias Error (NMBE) and the Coefficient of Variation of the Root Mean Square Error (CVRMSE). NMBE measures the systematic bias (global accuracy), while CVRMSE measures the variance of the errors. We reconstruct the physical energy consumption values from the discrete class predictions using the expm1 function.
\vspace{-0.5em}
\begin{equation}
    \text{NMBE} = \frac{\sum_{i=1}^{N} (\hat{y}_i - y_i)}{\sum_{i=1}^{N} y_i} \times 100\%,
\end{equation}
\vspace{-0.8em}
\begin{equation}
    \text{CVRMSE} = \frac{\sqrt{\frac{1}{N} \sum_{i=1}^{N} (\hat{y}_i - y_i)^2}}{\bar{y}} \times 100\%,
\end{equation}
where $y_i$ is the ground truth total energy of the $i$-th tile, $\hat{y}_i$ is the predicted total energy, $\bar{y}$ is the mean of the ground truth values, and $N$ is the number of samples in the test set.



\subsection{Generative Performance Evaluation}\label{gpe}
In this section, after training our framework, we evaluate the quality of the three generated outputs: satellite imagery, building height maps, and building energy consumption maps. 

\subsubsection{Quantitative Evaluation}

 We perform a quantitative image assessment across three fundamental dimensions: fidelity, diversity, and precision. The framework achieves a Peak Signal-to-Noise Ratio (PSNR) of 14.5, indicating that the generative distribution effectively simulates the statistical properties of real urban satellite imagery. A Learned Perceptual Image Patch Similarity (LPIPS) score of 0.430, complemented by diversity metrics such as SSIM (0.234) and FSIM (0.669), confirms the ability to generate diverse urban images. Moreover, precision assessments (an MS-SSIM of 0.240, SSIM of 0.182, and FSIM of 0.660) validate the high structural fidelity and visual alignment of the generated urban imagery.  In summary, these metrics demonstrate that our framework with Cotrolnet successfully encodes complex urban morphologies, generating compliant, realistic, and diverse urban imagery.

We further evaluate the performance of the H-Decoder (Height) and E-Decoder (Energy) using standard segmentation metrics \cite{chen2017rethinking} in Fig. \ref{fig:cm} and Tab. \ref{tab:segmentation_results} (see Appendix section~\ref{SHE}).

\begin{figure}[h]
    \vspace{-1em}
    \includegraphics[width=1.1\linewidth]{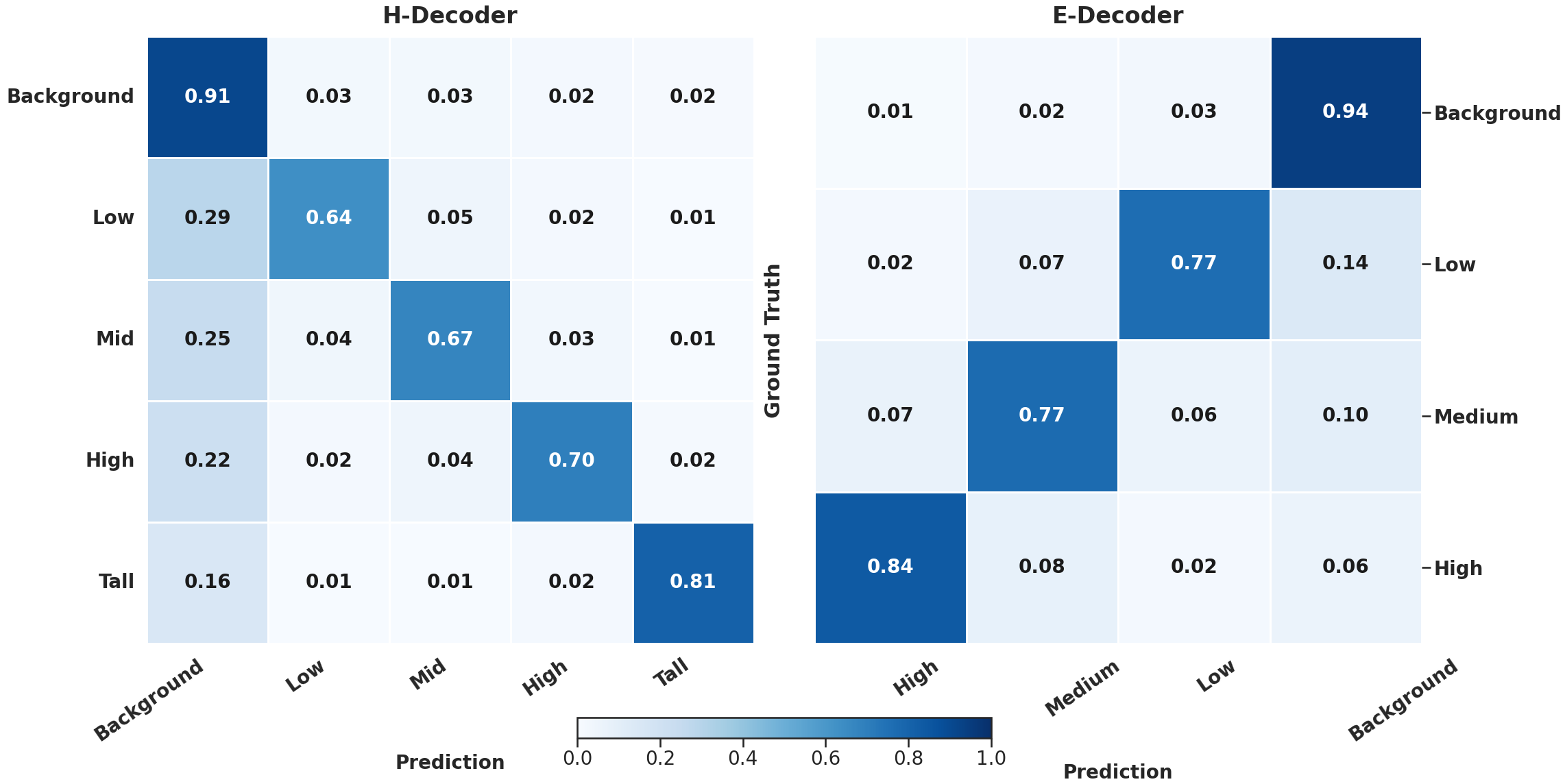}
    \vspace{-2em}
    \caption{Normalized confusion matrix for H-Decoder and E-Decoder.}
    \label{fig:cm}
    \vspace{-1em}
\end{figure}

The performance of H-Decoder reveals some clear patterns. We find: (1) The H-Decoder achieves an overall accuracy of 85.75\% and an mIoU of 0.6005. (2) Class 0 (Non-building) achieves the highest IoU (0.8443), demonstrating our framework's effectiveness in generating building footprints. (3) Class 4 (tall buildings) achieves a high IoU of 0.6664 and Recall of 0.8099, suggesting the latent features for tall buildings (e.g., shadows, large roof areas) are highly distinctive. (4) The confusion matrix in Fig. \ref{fig:cm} exhibits a good diagonal dominance. On the other hand, the E-Decoder demonstrates the capability of our framework to extract invisible energy patterns from visual latents, achieving an overall accuracy of 89.25\% and an mIoU of 0.7093. A key finding is the model's sensitivity to high-energy consumption. In Fig. \ref{fig:cm} and Tab. \ref{tab:segmentation_results}, Class 3 (high-energy consumption) yields a significantly higher performance compared to low-energy consumption. It is important for urban planning to identify high-energy-consumption buildings in UBEM.


\begin{figure}[h]
    \centering
    \vspace{-1em}
    \includegraphics[width=1\linewidth]{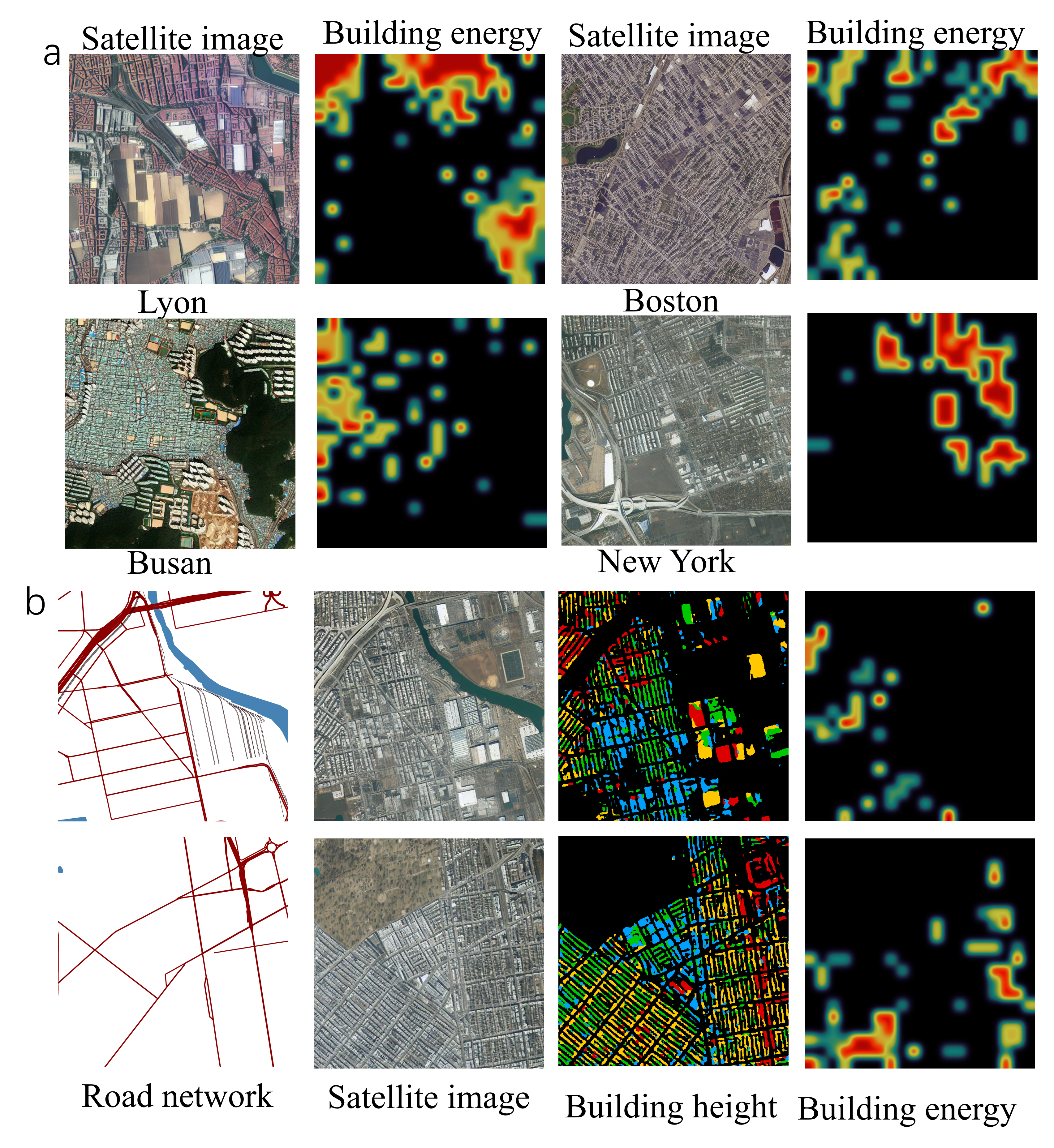}
    \vspace{-2em}
\caption{\textbf{Qualitative visualization of generated results across diverse cities.} (a) The generated building energy consumption maps and generated satellite imagery. (b) The generated satellite imagery, building height and energy consumption maps conditioning on road networks and prompts.}
    \label{fig:qualitative_results}
    \vspace{-1em}
\end{figure}

\begin{figure}[h]
\includegraphics[width=1\linewidth]{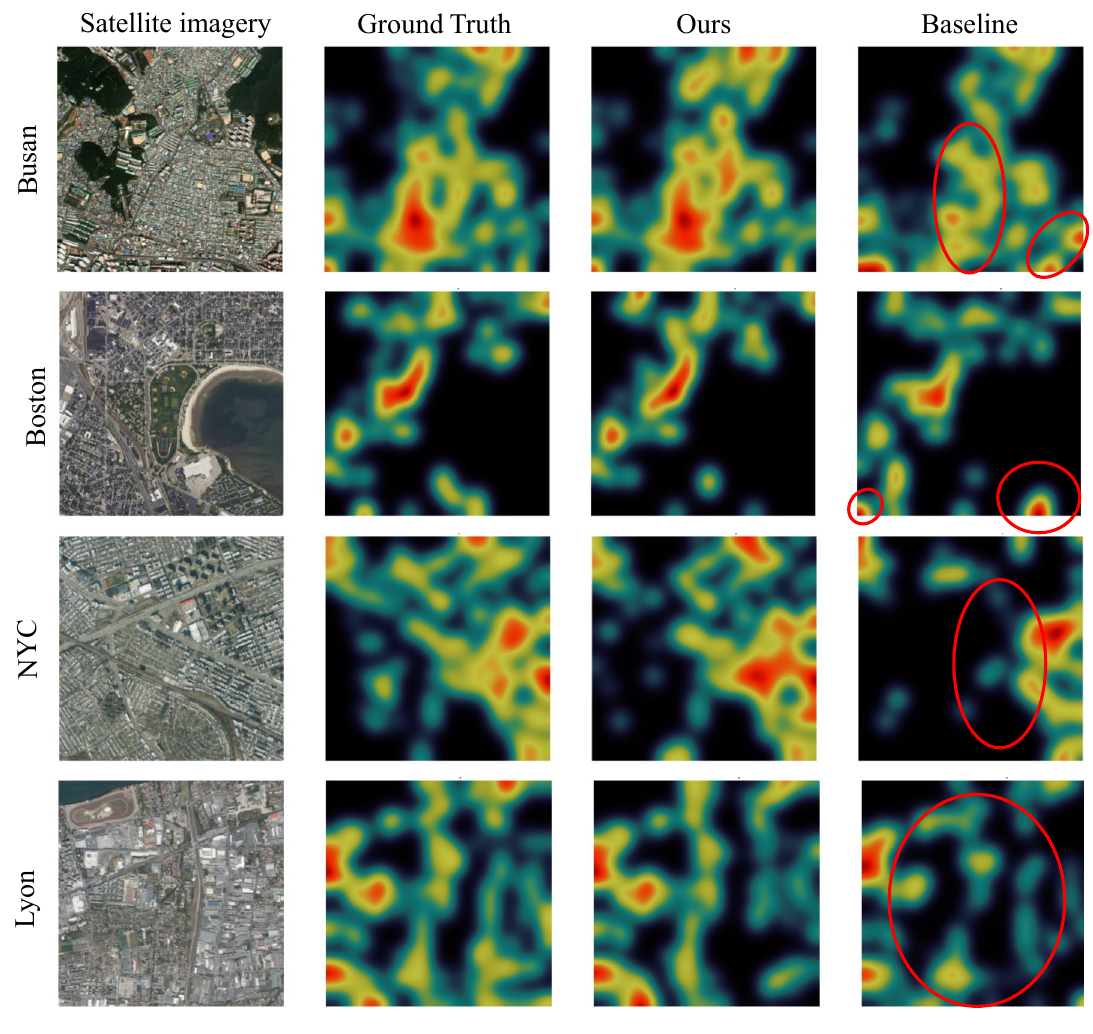}
\vspace{-2em}
\caption{\textbf{Performance comparison of building energy consumption prediction across different methods. Red cycles represent prediction errors by baseline \citet{yap2025revealing}. Baseline predicts more (e.g., in Busan, Boston) or less energy consumption (e.g., in NYC, Lyon, Busan) than ground truth.}}
\label{fig:Visualization}
\vspace{-2em}
\end{figure}

\subsubsection{Qualitative Evaluation}

Fig. \ref{fig:qualitative_results} provides the visualization of the generated samples across four distinct metropolitan areas: NYC, Boston, Lyon, and Busan. In Fig. \ref{fig:qualitative_results} (a), our framework successfully captures the unique urban morphology of each city. For NYC, the model reproduces the grid structures and high-density buildings typical of Manhattan. In contrast, for Lyon and Busan, it accurately synthesizes the irregular, organic street patterns and complex winding road networks inherent to historical European and mountainous Asian cities, respectively. A critical observation is the accurate spatial alignment between satellite imagery and generated building energy consumption maps. For instance, in the Lyon and New York samples, large-footprint institutional buildings are correctly highlighted as high-energy-consumption areas (red in Fig. \ref{fig:qualitative_results} (a)), while surrounding low-density residential areas are mapped to buildings with lower energy consumption (Busan and Boston samples). In Fig. \ref{fig:qualitative_results} (b), we show some generated satellite imagery, building height and energy consumption maps conditioned on road networks and text prompts. We find that our generated satellite imagery is spatially aligned with the road network. In summary, these results demonstrate the effectiveness of our GenAI framework to generate realistic, diverse and spatially aligned urban scenes (imagery, building height and energy consumption).

\begin{figure*}[h]
    \includegraphics[width=0.88\linewidth]{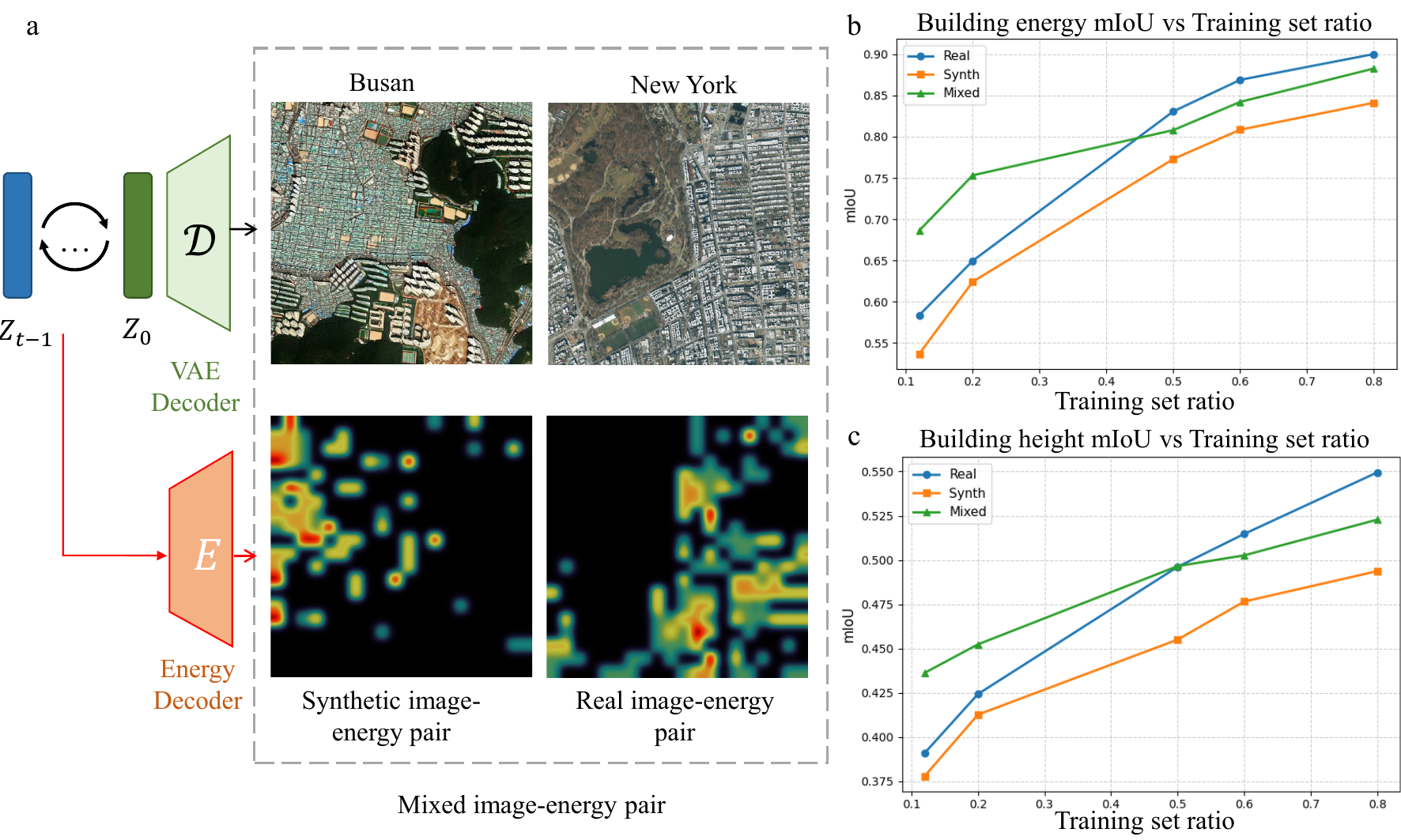}
    \vspace{-1em}
    \caption{\textbf{Performance comparison of downstream building energy consumption and height prediction tasks under different training data.} The x-axis represents the percentage of real data used. The y-axis denotes the mIoU on the fixed real test set. (b) Building energy consumption prediction results. (c) Building height prediction results.}
    \label{fig:downstream_charts}
    \vspace{-1em}
\end{figure*}

\subsubsection{Energy Performance Evaluation}
Beyond the performance evaluation through computer vision metrics, Tab. \ref{tab:physical_metrics} shows the comparison between our framework and standard calibration criteria. We evaluate the NMBE and CVRMSE using the expm1 function. Remarkably, our framework achieves an NMBE of 3.05\%. This result falls well within the strict calibration tolerance ($\pm 10\%$) typically required for energy models. Regarding variance, the model's CVRMSE is 14.62\%, which belongs to the standard range of 30\%.  It confirms SENSE can capture the physical energy patterns of the urban environment and achieve good performance. Note that our GenAI framework infers and generates energy value only from the latent features of a generated satellite image (without explicit information of weather, building material, type, HVAC, etc.).

\vspace{-1em}
\begin{table}[htbp]
    \centering
    \caption{Physical energy performance evaluation.}
    \vspace{-1em}
    \label{tab:physical_metrics}
    \begin{tabular}{lcc}
        \toprule
        \textbf{Metric}  & Ours & ASHRAE guideline 14 \textsuperscript{a} \\
        \midrule
        \textbf{NMBE (\%)}  & +3.05\% & $\pm 10\%$ \\
        \textbf{CVRMSE (\%)} & 14.62\% & 30\% \\
        \bottomrule
    \end{tabular}
    \par\smallskip
    \raggedright
    \footnotesize{\textsuperscript{a} ASHRAE Guide 14 considers a building model calibrated if hourly MBE values fall within ±10\% \cite{american2014ashrae,ROYAPOOR2015109}.}
    \vspace{-1.5em}
\end{table}

      

\begin{table*}[!htbp]
    \centering
    \fontsize{5}{2}\selectfont
    \caption{Energy prediction performance comparison. All models were fairly evaluated on a fixed real test set (20\%).}
    \vspace{-1em}
    \label{tab:energy-overall-energy}
    \setlength{\tabcolsep}{5.5pt}
    \resizebox{0.95\textwidth}{!}{
    \begin{tabular}{lcc cccc}
        \toprule
        Method & Acc$\uparrow$ & mIoU$\uparrow$ & NMBE & CVRMSE & NMBE (log) & CVRMSE (log) \\
        \midrule
        \citet{streltsov2020estimating} & 0.8277 & 0.5705 & -7.40\% & 18.84\% & -1.65\% & 5.20\% \\
        \citet{streltsov2020estimating} + \textbf{Ours} & \textbf{0.8766} & \textbf{0.6746} & \textbf{-1.32\%} & \textbf{15.96\%} & \textbf{-0.72\%} & \textbf{3.53\%} \\
        \text{Diff } ( |\text{Ours}| - |\text{Baseline}| )& +0.0489 & +0.1041 & -6.08\% & -2.88\% & -0.93\% & -1.67\% \\
        \hline
        \citet{xie2021segformer} & 0.8322 & 0.5791 & 7.53\% & 18.24\%& -1.13\% & 6.93\% \\
        \citet{xie2021segformer} + \textbf{Ours} & \textbf{0.8734} & \textbf{0.6845} & \textbf{-4.43\%}&   \textbf{17.23}\%& \textbf{-0.23\%} & \textbf{4.48\%} \\
        \text{Diff } ( |\text{Ours}| - |\text{Baseline}| ) & +0.0412 & +0.1054 & -3.1\% & -1.01\% & -0.90\% & -2.45\% \\
        \hline
        \citet{MAYER2023120542} & 0.8059 & 0.5380 & -12.66\% & 29.57\% & -2.34\% & 6.46\% \\
        \citet{MAYER2023120542} + \textbf{Ours} & \textbf{0.8321} & \textbf{0.5926} & \textbf{1.58\%} & \textbf{23.73\%} & \textbf{-0.86\%} & \textbf{4.16\%} \\
        \text{Diff } ( |\text{Ours}| - |\text{Baseline}| )  & +0.0262 & +0.0546 & -11.08\% & -5.84\% & -1.48\% & -2.30\% \\
        \hline
        \citet{yap2025revealing} & 0.8692 & 0.6493 & -16.49\% & 24.91\% & -3.20\% & 6.72\% \\
        \citet{yap2025revealing} + \textbf{Ours} & \textbf{0.9118} & \textbf{0.7581} & \textbf{-5.03\%} & \textbf{14.99\%} & \textbf{-1.19\%} & \textbf{4.03\%} \\
         \text{Diff } ( |\text{Ours}| - |\text{Baseline}| )  & +0.0426 & +0.1088 & -11.46\% & -9.92\% & -2.01\% & -2.69\% \\
        \bottomrule
    \end{tabular}
    }
    \vspace{-1.5em}
\end{table*}

\begin{table*}[htbp]
    \centering
    \caption{Segmentation performance comparison for buildings with Background, Low-, Medium-, and High-energy consumption.}
    \vspace{-0.8em}
    \label{tab:segmentation-performance}
    \setlength{\tabcolsep}{4.5pt}
    \resizebox{0.98\textwidth}{!}{
    \begin{tabular}{l cccc cccc cccc}
        \toprule
        \multirow{2}{*}{Method} &
        \multicolumn{4}{c}{IoU (per class)} &
        \multicolumn{4}{c}{Precision (per class)} &
        \multicolumn{4}{c}{Recall (per class)} \\
        \cmidrule(lr){2-5}\cmidrule(lr){6-9}\cmidrule(lr){10-13}
        & Background & Low & Medium & High
        & Background & Low & Medium & High
        & Background & Low & Medium & High \\
        \midrule
        \citet{streltsov2020estimating} &
        0.8369 & 0.4280 & 0.4656 & 0.5514 &
        0.8961 & 0.5986 & 0.6885 & 0.7389 &
        0.9269 & 0.6003 & 0.5898 & 0.6848 \\
        \citet{streltsov2020estimating} + \textbf{Ours} &
        \textbf{0.8782} & \textbf{0.5808} & \textbf{0.5892} & \textbf{0.6504} &
        \textbf{0.9307} & \textbf{0.7561} & \textbf{0.7385} & \textbf{0.7935} &
        \textbf{0.9396} & \textbf{0.7147} & \textbf{0.7445} & \textbf{0.7829} \\
        Diff (Ours$-$Baseline) &
        +0.0413 & +0.1528 & +0.1236 & +0.0990 &
        +0.0346 & +0.1575 & +0.0500 & +0.0546 &
        +0.0127 & +0.1144 & +0.1547 & +0.0981 \\
        \midrule
        \citet{xie2021segformer} &
        0.8489 & 0.4452 & 0.4760 & 0.5464 &
        0.9173 & 0.6106 & 0.6474 & 0.7172 &
        0.9192 & 0.6217 & 0.6426 & 0.6965 \\
        \citet{xie2021segformer} + \textbf{Ours} &
        \textbf{0.8735} & \textbf{0.5692} & \textbf{0.6056} & \textbf{0.6897} &
        \textbf{0.9620} & \textbf{0.6604} & \textbf{0.7290} & \textbf{0.7878} &
        \textbf{0.9047} & \textbf{0.8048} & \textbf{0.7814} & \textbf{0.8471} \\
        Diff (Ours$-$Baseline) &
        +0.0246 & +0.1240 & +0.1296 & +0.1433 &
        +0.0447 & +0.0498 & +0.0816 & +0.0706 &
        -0.0145 & +0.1831 & +0.1388 & +0.1506 \\
        \midrule
        \citet{MAYER2023120542} &
        0.8126 & 0.3944 & 0.4211 & 0.5240 &
        0.8888 & 0.6291 & 0.5747 & 0.6824 &
        0.9045 & 0.5139 & 0.6117 & 0.6930 \\
        \citet{MAYER2023120542} + \textbf{Ours} &
        \textbf{0.8319} & \textbf{0.4989} & \textbf{0.4614} & \textbf{0.5781} &
        \textbf{0.9118} & \textbf{0.6432} & \textbf{0.7001} & \textbf{0.6697} &
        \textbf{0.9046} & \textbf{0.6898} & \textbf{0.5751} & \textbf{0.8087} \\
        Diff (Ours$-$Baseline) &
        +0.0193 & +0.1045 & +0.0403 & +0.0541 &
        +0.0230 & +0.0141 & +0.1254 & -0.0127 &
        +0.0001 & +0.1759 & -0.0366 & +0.1157 \\
        \midrule
        \citet{yap2025revealing} &
        0.8675 & 0.5324 & 0.5618 & 0.6357 &
        0.8945 & 0.7797 & 0.7797 & 0.8541 &
        0.9663 & 0.6266 & 0.6678 & 0.7131 \\
        \citet{yap2025revealing} + \textbf{Ours} &
        \textbf{0.9081} & \textbf{0.6877} & \textbf{0.6819} & \textbf{0.7548} &
        \textbf{0.9442} & \textbf{0.8654} & \textbf{0.7911} & \textbf{0.8836} &
        \textbf{0.9596} & \textbf{0.7700} & \textbf{0.8317} & \textbf{0.8382} \\
        Diff (Ours$-$Baseline) &
        +0.0406 & +0.1553 & +0.1201 & +0.1191 &
        +0.0497 & +0.0857 & +0.0114 & +0.0295 &
        -0.0067 & +0.1434 & +0.1639 & +0.1251 \\
        \bottomrule
    \end{tabular}
    }
    \vspace{-1em}
\end{table*}

\vspace{-1em}
\subsection{Urban Data Augmentation Performance}

In Section \ref{gpe}, the framework was trained on all training samples. However, in practice, ground truth energy data is scarce and limited. To answer RQ3 and RQ4, we conducted data augmentation experiments on downstream urban building energy consumption and height tasks. 




We employ three different training strategies: (i) \textbf{Real Only (Baseline):} The prediction model is trained strictly on the limited available real data. (ii) \textbf{Mixed Training (ours):} The prediction model is trained on a combination of the limited real data and a large corpus of synthetic data (see Fig. \ref{fig:downstream_charts} a). (iii) \textbf{Synthetic Only:} The prediction model is trained exclusively on synthetic data generated by our framework and tested on real data. 

\subsubsection{Energy prediction performance comparison}

In Tab. \ref{tab:energy-overall-energy}, we compare different state-of-the-art methods on MUSE. To avoid data leakage, we train our generative framework (including two building decoders) and the downstream prediction models using the available 20\% real training data. All models were evaluated on another fixed real-world test set to ensure fair comparison.  Tab. \ref{tab:energy-overall-energy} reveals some clear findings: (1) The integration of our generative framework consistently enhances the performance of all baseline models, with mean Intersection over Union (mIoU) improving by 5.46\% to 10.88\% across all baselines. (2) The \citet{yap2025revealing} model achieves the highest performance gain (about an mIoU of 0.11) by data augmentation. (3) Furthermore, the NMBE shows marked improvement (e.g., the NMBE for \citet{streltsov2020estimating} is reduced from -7.40\% to -1.32\%), while the CVRMSE drops significantly (e.g., the CVRMSE for \citet{yap2025revealing} is reduced from 24.91\% to 14.99\%). 

Detailed analysis in Tab. \ref{tab:segmentation-performance} confirms that these improvements are consistent across all energy levels, with IoU improvements often exceeding 10\% in the Low, Medium, or High categories. These performance gains (e.g., the 15.75\% increase for Low-energy precision in the \citet{streltsov2020estimating} model) reveal that our framework can learning valid physical energy consumption patterns, serving as an effective data generator for limited real energy consumption data. The visualization results can be found at Fig. \ref{fig:Visualization}. In summary, our proposed framework improves existing state-of-the-art urban building energy prediction models with limited real data.


\subsubsection{Results on building energy consumption prediction}
To further analyse the impact of our generative framework on few-shot urban building prediction tasks. We train the same model using the available real training data to varying fractions (e.g., 10\%, 20\%, ..., 80\%), and test the model on a fixed test set (20\% MUSE). Fig. \ref{fig:downstream_charts} (b) illustrates the mIoU performance for building energy consumption prediction across different data settings. We find: (1) Increased training data improves building energy consumption prediction performance, regardless of different training strategies. (2) Although purely synthetic training data leads to a slight performance drop, the performance (orange line) remains competitive. It suggests that our E-Decoder has successfully learned the urban building energy distribution that generalizes well to real-world data. (3) Mixed data training yields a big gain (about 10\%)under limited data, showing strong benefits of synthetic augmentation. (4) The generated image and building energy consumption map are spatially aligned.

\vspace{-1em}
\subsubsection{Results on building height prediction}
Fig. \ref{fig:downstream_charts} (c) presents the results for building height prediction.  We find: (1) Increased training data improves height prediction performance, regardless of different training strategies. (2) Purely synthetic training data leads to a slight performance drop. (3) Mixed data training yields a big gain (4\%-7\%) under limited data, showing strong benefits of synthetic data augmentation and good generalization. (4) The generated image and height map are spatially aligned.

In summary, in Fig. \ref{fig:downstream_charts} (b-c), the "Mixed" training strategy (green line) consistently outperforms the "Real Only" baseline (Blue line) in few-shot settings, demonstrating the efficacy of our data augmentation, particularly when real data is scarce ($<40\%$).

\vspace{-1.2em}
\section{Discussion}

\subsection{Domain Contributions and Impacts}

As rapid urbanisation becomes an essential phenomenon across all global cities, optimizing the urban and building designs and improving the energy efficiency can make a significant contribution to UNs' SDGs 7 and 11. This study supports the integration of building energy consumption maps generation and analysis into urban science and planning. Traditionally, energy modeling is treated as a post-design assessment step. In contrast, our GenAI framework allows planners to use energy consumption as an active design parameter. By adjusting input constraints—such as road networks, building coverage ratios, and volume densities, policymakers can simulate various development scenarios and immediately visualize the resulting energy maps. Furthermore, policymakers can efficiently identify specific building typologies for energy efficiency retrofits, optimizing the allocation of decarbonization resources.

Besides, a major challenge for global decarbonization is the lack of comprehensive building energy records in developing and rapidly urbanizing cities. We only use less than 20\% manually labelled energy data to train SENSE, enabling virtually infinite annotated data generation. The results show that mixing synthetic data with a limited amount of real-world data (e.g., 20\%) significantly improves the accuracy of existing state-of-the-art urban building energy prediction models. It confirms that our generated building height and energy consumption maps contain valid semantic signals that help the downstream prediction networks generalize better on unseen real data. Prediction models achieve good performance with synthetic-only and mixed training strategies, verifying the accurate spatial alignment between satellite imagery and the generated building energy consumption or height map. Our findings align with the opinion \cite{wu2023datasetdm}: GenAI can act as infinite data generators. Hence, municipalities with limited data infrastructure can leverage our framework using publicly available satellite imagery, avoiding the high costs and time requirements of on-site data collection. 

Our framework offers a powerful tool for government agencies, urban and energy scientists, and developers to envision urban scenarios and conduct comparative analyses, particularly in data-scarce developing regions. This study reveals the capacity of large GenAI models to bridge the gap between planning concepts/constraints (e.g., road networks and density metrics) and concrete urban form, effectively transforming user-defined parameters into detailed, spatially aligned satellite imagery, energy, and height maps. Moreover, for downstream applications such as building energy consumption prediction, our findings confirm the efficacy of synthetic data in augmenting real training datasets to significantly boost prediction performance for sustainable development. By releasing our framework as an open-access tool, we advocate for a democratized planning process, empowering policymakers and communities alike to customize energy-effiency urban planning that actively aligns with SDGs 7 and 11.

\subsection{Limitations and Ethical Considerations}
This study has several limitations that represent opportunities for future research. Our study focuses on annual total energy consumption and does not capture temporal dynamics. Future work can incorporate temporal dimensions into the generative process to enable the synthesis of hourly energy profiles, even though it is hard to obtain this fine-grained energy data. While the model shows good generalisation across four big metropolitan areas, future work should expand the diversity and global coverage of the dataset.

All training data in MUSE are derived from publicly available sources, including OpenStreetMap, Mapbox, and municipal energy disclosure records (e.g., Local Law 84), with no use of private or non-consensual household data. As confidentiality often limits the sharing of fine-grained urban energy information, our framework mitigates this constraint by generating synthetic datasets with energy consumption annotations that spatially align with real-world urban environments. This enables high-quality data sharing and algorithm benchmarking while avoiding privacy issues, facilitating scientific discoveries and technical advancements in AI for Science.

\section{Conclusion}
This study proposes a unified multi-modal GenAI framework that synthesizes spatially aligned satellite imagery, building height, and energy consumption maps. It also establishes a global Multi-city Urban Satellite-Energy Dataset (MUSE). Experiments demonstrate that urban building energy consumption and height can be reliably generated from the latent space in existing GenAI models. For the limited physical energy data issue, experiments demonstrate that our framework can improve the performance of existing SOTA physical energy prediction models, significantly reducing energy prediction error with 3\%-11\% NMBE and 1\%-9\% CVRMSE. Our dataset and framework provide a foundation for AI-driven scientific discovery across urban, energy, and building sciences.



\clearpage
\section{GenAI Disclosure}

The authors declare that artificial intelligence (AI) tools, specifically Gemini, were used solely to assist with language polishing and grammar checking of the manuscript text. All intellectual content, data analysis, interpretations, and conclusions were conceived, written, and verified by the authors.

\bibliographystyle{ACM-Reference-Format}
\bibliography{sample-base}

\appendix

\section{Appendix}
\subsection{Implementation Details}
\label{sec:Implementation}

\subsubsection{Building Height Data}
Given the inherent noise in satellite-derived estimates and the high variance in urban data, we formulate the building height co-generation tasks as segmentation problems rather than continuous regression. This requires discretizing the continuous ground truth values into categorical labels. To ensure statistical validity and alleviate class imbalance during training, we employed a quantile-based discretization strategy. In particular, the building heights are categorized into 5 classes. Class 0 represents non-building background pixels. For building pixels, we calculate the $25^{th}$, $50^{th}$, and $75^{th}$ percentiles ($Q_1, Q_2, Q_3$) of the height distribution. The classes are assigned as follows: Class 1 (Low-rise, $h < Q_1$), Class 2 (Mid-rise, $Q_1 \leq h < Q_2$), Class 3 (High-rise, $Q_2 \leq h < Q_3$), and Class 4 (Super-tall, $h \geq Q_3$). 

\subsubsection{Building Energy Data}
Energy consumption is divided into 4 classes. Class 0 denotes non-energy background. The remaining building pixels are divided into 3 intervals based on the tertiles ($33^{rd}$ and $66^{th}$ percentiles) of the logarithmic energy consumption distribution. This results in Class 1 (Low Energy), Class 2 (Medium Energy), and Class 3 (High Energy). Log-transformation was applied prior to discretization to handle the long-tail distribution typical of urban energy consumption data.

\subsubsection{Hyperparameters}
Our framework includes ControlNet, which is finetuned with high-resolution satellite imagery at $512 \times 512$ pixels. To maintain maximum numerical stability and model performance, we utilize full FP32 precision throughout the training phase. The training is conducted with a batch size of 16, leveraging the Distributed Data Parallel (DDP) strategy to ensure efficient and synchronized gradient updates across multiple computation nodes. Under this DDP configuration, each GPU maintains a complete replica of the model and processes unique subsets of the training data. Gradients are synchronized during each backward pass through high-performance NCCL-based communication. The pretrained weights for ControlNet and standard Stable Diffusion (version 1.5) were sourced from the official repository provided by \footnote{https://github.com/lllyasviel/ControlNet}.

Our framework includes H-Decoder and E-Decoder. In this study, the H-Decoder is implemented using a SegFormer architecture with a MiT-B3 backbone at a resolution of 512 $\times$ 512 pixels using the AdamW optimizer with an initial learning rate of $3 \times 10^{-4}$. We utilize a batch size of 16 and apply a composite loss function comprising Cross-Entropy and Dice loss weighted by class frequency. Furthermore, the training process incorporates data augmentation techniques, including random flips and 90-degree rotations. Similarly, E-Decoder is trained for 30 epochs with a batch size of 16 using the AdamW optimizer ($lr = 3 \times 10^{-4}$, weight decay =$ 1 \times 10^{-4}$), employing a composite loss function comprising Cross-Entropy and Dice loss weighted by class frequency.

\subsection{Segmentation performance for H-Decoder and E-decoder}\label{SHE}

\begin{table*}[h]
\centering
\small
\caption{Class-wise segmentation performance for H-Decoder and E-decoder. The model effectively captures extreme values (Background and High-consumption/Tall buildings).}
\label{tab:segmentation_results}
\begin{tabular*}{\textwidth}{@{\extracolsep{\fill}}lcccccc}
\toprule
\textbf{Task} & \textbf{Class ID} & \textbf{Description} & \textbf{Precision} & \textbf{Recall} & \textbf{IoU} & \textbf{Dice} \\
\midrule
\multirow{5}{*}{\textbf{Building Height}} 
& 0 & Background & 0.9182 & 0.9129 & 0.8443 & 0.9156 \\
& 1 & Low-rise & 0.6548 & 0.6362 & 0.4764 & 0.6454 \\
& 2 & Mid-rise & 0.6402 & 0.6721 & 0.4878 & 0.6558 \\
& 3 & High-rise & 0.6832 & 0.6983 & 0.5275 & 0.6906 \\
& 4 & Tall & 0.7899 & 0.8099 & 0.6664 & 0.7998 \\
\cmidrule{2-7}
\textbf{Overall (Avg)} & - & \textbf{0.8575 (Accuracy)}
 &0.7373 & 0.7459 & 0.6005 & 0.7414 \\
\midrule
\midrule
\multirow{4}{*}{\textbf{Building Energy}} 
& 0 & Background & 0.950 & 0.942 & 0.898 & 0.946 \\
& 1 & Low consumption & 0.756 & 0.765 & 0.613 & 0.760 \\
& 2 & Medium consumption & 0.764 & 0.773 & 0.624 & 0.769 \\
& 3 & High consumption & 0.812 & 0.837 & 0.702 & 0.825 \\
\cmidrule{2-7}
\textbf{Overall (Avg)} & - & \textbf{0.8925 (Accuracy)} & 0.8205 & 0.8293 & 0.7093 & 0.8250 \\
\bottomrule
\end{tabular*}
\end{table*}
      
\subsection{Data filtering}\label{data-filter}
\begin{figure*}[h]
    \centering
    \includegraphics[width=0.9\linewidth]{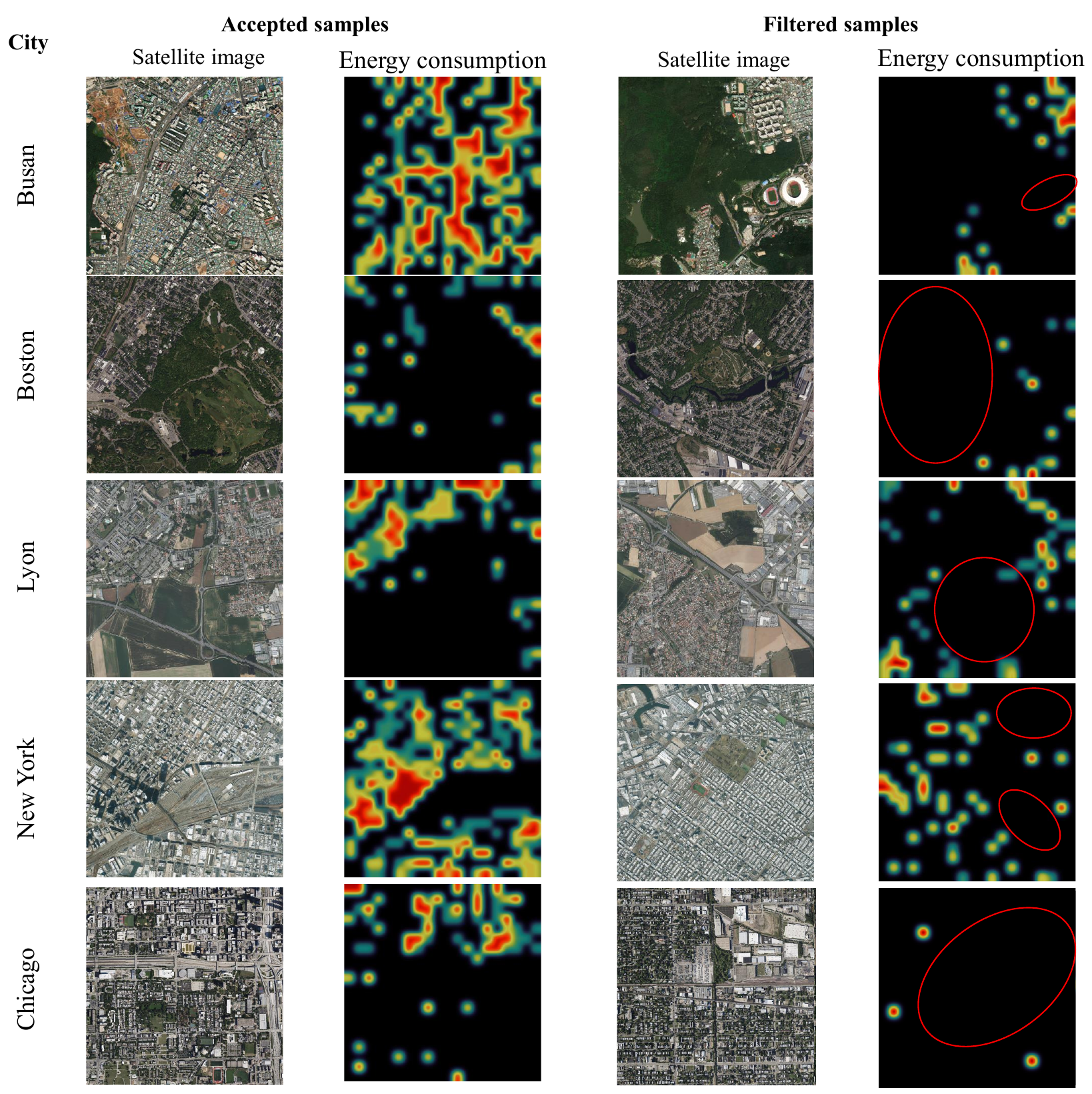}
    \caption{Data filtering. We filtered out samples that clearly lacked building energy labels.}
    \label{fig:datafilter}
\end{figure*}

\begin{table*}
\centering
\caption{City-level statistics data in MUSE.}
\vspace{-1em}
\label{tab:city_dataset_stats}
\setlength{\tabcolsep}{6pt}
\begin{tabular}{lccccc}
\toprule
City & Satellite imagery &Geospatial constraint & Urban density &Original energy map  & Filtered energy map  \\
\midrule
New York City & 1589& 1589& 1589 & 1589& 579 
 \\
Boston & 2051& 2051& 2051 & 2051& 526 
 \\
Lyon & 1412& 1412& 1412 & 1412& 687 
 \\
Busan & 1438& 1438& 1438 & 1438& 996  \\
\bottomrule
\vspace{-1em}
\end{tabular}
\end{table*}

\end{document}